# Impact of Video Compression on the Performance of Object Detection Systems for Surveillance Applications

Michael O'Byrne, Mark Sugrue
Kinesense Ltd.
79 Merrion Square,
Dublin 2, Ireland.
{m.obyrne, mark}@kinesensevca.com

Vibhoothi, Anil Kokaram
Sigmedia Group, Department of Electronic and
Electrical Engineering,
Trinity College Dublin, Dublin, Ireland.
{vibhoothi, Anil.Kokaram}@tcd.ie

## Abstract

*This study examines the relationship between H.264 video compression and the performance of an object detection network (YOLOv5). We curated a set of 50 surveillance videos and annotated targets of interest (people, bikes, and vehicles). Videos were encoded at 5 quality levels using Constant Rate Factor (CRF) values in the set {22,32,37,42,47}. YOLOv5 was applied to compressed videos and detection performance was analyzed at each CRF level. Test results indicate that the detection performance is generally robust to moderate levels of compression; using a CRF value of 37 instead of 22 leads to significantly reduced bitrates/file sizes without adversely affecting detection performance. However, detection performance degrades appreciably at higher compression levels, especially in complex scenes with poor lighting and fast-moving targets. Finally, retraining YOLOv5 on compressed imagery gives up to a 1% improvement in $F_1$ score when applied to highly compressed footage.*

## 1. Introduction

The security industry is continuously pushing for enhanced surveillance systems capable of high-resolution monitoring. For such systems, lossy video compression is essential to reduce the massive quantities of raw data before transmission. At the same time, there is growing interest in the use of video analytics to automatically interpret the visual data. For surveillance systems that leverage video analytics, it is crucial that a suitable video compression rate is chosen such that the quality of the compressed video is sufficient for downstream analysis tasks. However, system configuration decisions are often made without considering specific task requirements, such as the quality needed for reliable object detection.

While reducing the video resolution is another way to reduce bitrate, surveillance system operators will typically want to retain high-resolution videos for occasional visual inspection and archival purposes. For that reason, video compression using an effective codec is often the preferred way to control bitrates/file sizes.

The effect of video compression on video quality is an important practical consideration that is often overlooked in the design of surveillance monitoring systems. This paper investigates the trade-off between video compression using the H.264 standard and the performance of a popular object detection network, namely, YOLOv5 [1]. The YOLOv5 model is trained to detect three object categories: persons, bikes (includes bicycles and motorcycles) and vehicles.

We show that the object detection network can tolerate high levels of compression in certain scenarios, however, detection performance starts to breakdown when dealing with highly compressed imagery from scenes that are characterized by challenging illumination conditions. Moreover, the detection of small, low-contrast, and fast-moving objects is particularly hampered at higher compression levels.

Finally, we investigate ways to improve the performance of object detection systems when applied to highly compressed imagery by retraining YOLOv5 with an additional corpus of corrupted images. Commonly, deep neural networks are trained on relatively good quality image datasets, yet in real-world settings, the input video footage has often been subjected to extensive compression and cannot be assumed to be of high quality. In an effort to boost detection performance when dealing with compressed videos, we created a training dataset of 22,571 degraded images via data augmentation (the source of the images is the MS COCO dataset [2]). These images feature realistic video compression artefacts and other characteristics of surveillance style imagery such as overlaid timestamps.

### 1.1. Related work

There are several works in the literature that examine the effects of lossy image compression on the performance of computer vision models. Poyser et al. [3] investigated the impact of H.264 compression on the performance of a human action recognition model and found that there was a

This work was funded by the Disruptive Technologies Innovation Fund, Project: 166687/RR. This paper is accepted in AVSS 2022, ©IEEE 2022.

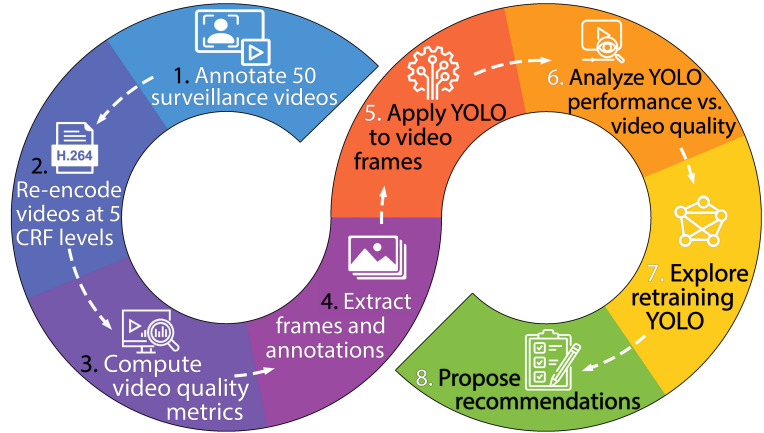

Figure 1: Key stages involved in the design and implementation of this study.

significant decrease in performance when using a CRF value above 40. They also investigated the impact of JPEG (Joint Photographic Experts Group) compression across four discrete tasks: human pose estimation, semantic segmentation, object detection, and monocular depth estimation. They found that models for each of these tasks could tolerate moderate levels of JPEG compression, however, there was a notable drop in performance when using a JPEG quality (quantization) level below 15%. Additionally, they found that retraining models on compressed imagery leads to a performance gain when models are applied to images that are similarly compressed. Likewise, Zanjani et al. [4] and Benbarrad et al. [5] found that employing compression-based data augmentation as part of the model training process is an effective strategy for improving the performance of classification models when applied to highly compressed imagery.

Dodge and Karam [6] and Roy et al. [7] explored the impact of several types of quality distortions (blur, noise, contrast, JPEG compression etc.) on the performance of image classification models. They both showed that classification models are generally resilient to all but the most severe levels of JPEG compression, however, they are more susceptible to blur and noise.

Gandor and Nalepa [8] considered the impact of image compression on the performance of nine off-the-shelf object detection models. They found that JPEG compression is generally friendly to object detectors, but unlike findings related to the influence of JPEG compression on image classification models, there was a more pronounced decline in detection performance with decreasing JPEG quality.

Klare and Burge [9] analyzed the impact of H.264 video compression on face recognition performance. They report a non-linear relationship between recognition performance and bitrate. Videos could be compressed down to 128kb/s before a notable drop in recognition performance occurred.

### 1.2. Contribution

Our test video dataset specifically targets surveillance applications. It contains a rich collection of real-world surveillance videos and covers a broad range of situations such as day/night scenes, indoor/outdoor scenes, different weather conditions, different types of footage (e.g., CCTV, dashcam, body-worn) etc. The comprehensive and diverse nature of our test dataset means that we can, at a more granular level, identify scenarios where the use of heavily compressed video leads to particularly poor detection performance. Moreover, while previous studies rely on off-the-shelf models that are intended for general-purpose applications and are trained to detect many non-relevant classes such as bananas, elephants, toothbrushes etc., we use a model tailored to surveillance applications that detects objects of core interest (i.e. persons, bikes, and vehicles). These factors increase the relevance and validity of our findings when applied to surveillance applications, and this represents an important practical contribution as having knowledge of the relationship between input compression and output detection performance can inform design decisions regarding future video devices and infrastructure.

In addition, this paper examines the relationship between input video quality, as measured using a variety of metrics, with the detection performance at each compression level. We also report on the extent to which performance degradation can be recovered when networks are applied to highly compressed surveillance footage by incorporating compressed images in the training dataset.

### 2. Methodology

The key steps in this study are highlighted in Figure 1. The first step involves sourcing and annotating the surveillance videos and is detailed in Section 2.1. This is followed by steps 2 and 3 which deal with systematically compressing the videos and computing video quality metrics, as discussed in Section 2.2. Step 4 concerns the process of generating the test datasets at each compression level, which is outlined in Section 2.3. Steps 5 and 6 relate to training the YOLOv5 model and evaluating detection performance, details of which are provided in Section 2.4. Step 7 looks at boosting detection performance by retraining the model with an additional corpus of compressed training data, as is described in Section 2.5. The final step consists of proposing recommendations based on the outcome of this study and this is done in Section 4.

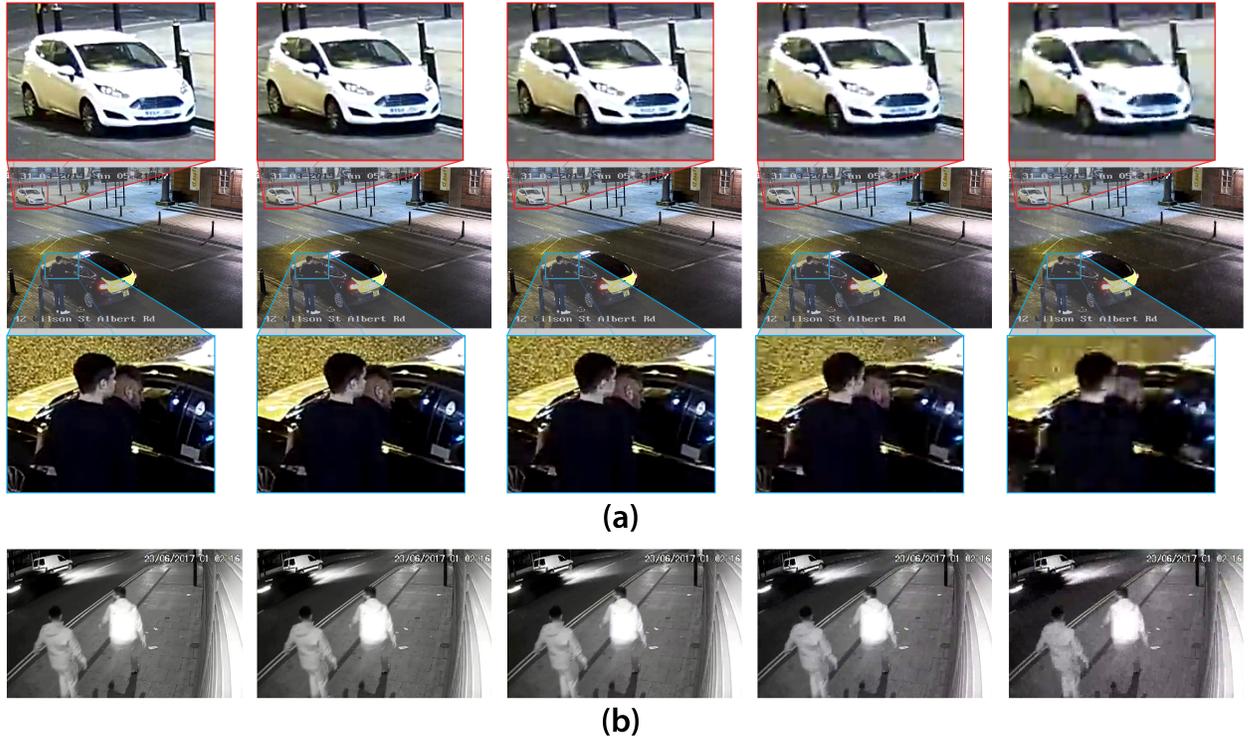

Figure 2: Sample videos from our test dataset. Left to Right: Frames from videos encoded with CRF values of 22,32,37,42,47. Top to bottom: (a) night-time street scene with close-up views of a stationary vehicle and persons. The vehicle holds up well under increasing compression, however, the persons become more ambiguous, and (b) a near-Infrared (nIR) scene with moving vehicle and persons.

### 2.1. Test dataset curation

The test dataset consists of 50 surveillance videos that were carefully selected such that they cover a wide range of illumination and weather conditions, as well as featuring day/night-time scenes, different video resolutions (mostly 360p, 480p, 720p, and 1080p), and various types of surveillance footage (e.g., CCTV, dashcam, and body-worn footage etc.). This diversity was necessary so that we could pinpoint scenarios where the use of high compression rates had a particularly adverse impact on detection performance.

A total of 5,790 frames were extracted from these test videos and objects of interest were labelled. There were 13,924, 1,633, and 18,695 bounding box annotations extracted for person, bike and vehicle classes, respectively.

### 2.2 Video compression and video quality metrics

Each of the test videos was encoded using the H.264 codec at 5 different Constant Rate Factor (CRF) levels (CRF = 22, 32, 37, 42, 47), which gave rise to 5 output videos for each test video. An example of two test videos encoded at each CRF level is shown is Figure 2. H.264 was chosen as it is a widely used codec in the CCTV industry. We have used the x264 implementation of the H.264 codec.

The bitrate and various video quality metrics were computed for each video. The quality metrics were PSNR (Peak Signal-to-Noise Ratio), SSIM (Structural Similarity Index) [10], and VMAF (Video Multi-Method Assessment Fusion) [11-12]. VMAF differs from the PSNR and SSIM scores insofar as it is a perceptual video quality metric that aims to approximate human perception of video quality. VMAF estimates the perceived quality score by computing scores from several quality assessment algorithms and fusing them using support vector machines (SVMs). VMAF scores range from 0 to 100, with 0 indicating the lowest quality, and 100 the highest. The goal of computing metrics was to see if they correlate with detection performance.

### 2.3 Extracting frames from compressed videos

Frames in the compressed videos that correspond with annotated frames in the original test videos were extracted,

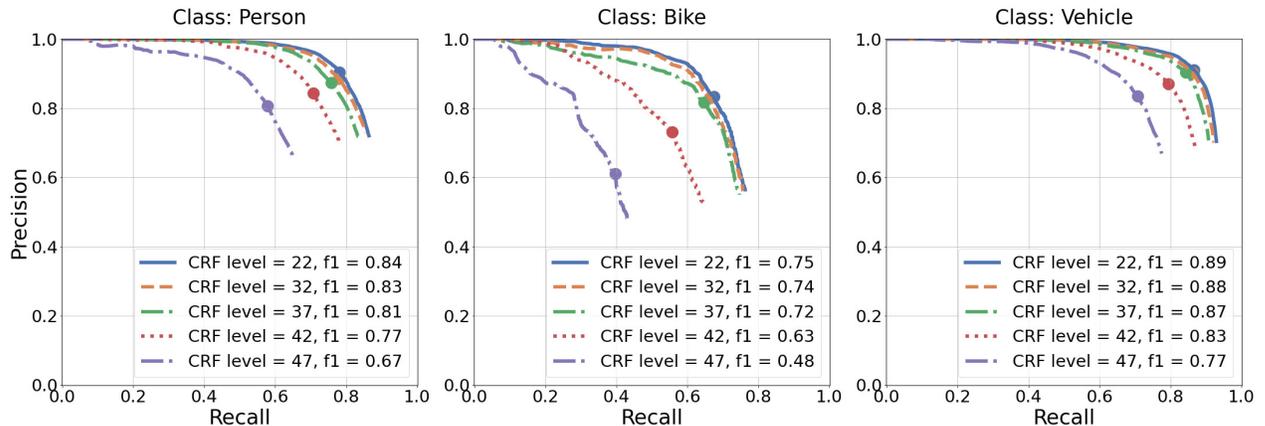

Figure 3: Precision-Recall Curves for YOLOv5 when applied to footage encoded at multiple CRF values. Left to right: PR curves for Person, Bike and Vehicle classes. We can see that YOLOv5 performs consistently at the low-moderate compression levels (CRF = 22, 32, 37), however, performance declines markedly at higher compression levels (CRF = 43,47), particularly for the bike class.

along with the corresponding PSNR, SSIM, and VMAF scores at each frame time.

### 2.4 Details of the YOLOv5 network

The object detection network was a custom YOLOv5 model that was designed to detect 3 object categories, namely: persons, bikes, and vehicles. The model was trained on a dataset of 229,489 images that is a combination of several open-source datasets as well as our own in-house dataset featuring surveillance style footage. The input image size into the network was 544x544 pixels.

YOLOv5 was applied to the extracted frames at each CRF level, and an in-depth analysis was subsequently carried out, which is presented in Section 3.

### 2.5 Retraining YOLOv5 with compressed imagery

Representative training data is the bedrock for creating reliable and effective deep neural networks. The vast majority of imagery in open-source datasets is not representative of surveillance style footage. This is particularly pertinent when dealing with highly compressed video streams containing prominent compression and motion artefacts, which may mislead deep learning methods that were trained on good quality, minimally compressed images. To address this, we retrain YOLOv5 using an additional corpus of compressed training images that mimic the compression artefacts seen on compressed surveillance video data.

### 3. Results

The performance of the YOLOv5 network for each object category is shown in the Precision-Recall (PR) curves in Figure 3. It may be observed that the detection performance is consistent for the three lowest compression levels (CRF = 22,32,37). The $F_1$ scores for each object category differ by less than 3% when a using a CRF value of 22 versus a CRF value of 37. At higher compression rates (CRF = 42), we begin to see a notable drop in the detection performance, which is further exacerbated at the highest compression setting (CRF = 47), with the bike class suffering the greatest performance degradation. One possible explanation for this is that bikes appear as fine-structured objects and these slender bodies can be easily destroyed during the compression process. In contrast, vehicles typically appear as larger and more substantive bodies and can retain their general shape even when subjected to extensive compression, as seen in Figure 2(a).

### 3.1. Effect of object size on detection success

The size and detection status of objects are illustrated in Figure 4. It may be noted that larger objects tend to be successfully detected in both minimally compressed (CRF = 22) and highly compressed (CRF = 47) cases. However, smaller objects are more susceptible to the effects of video compression as many of these objects that are detectable at CRF = 22 are not detectable at CRF = 47. We can see this clearly in the case of the person class. With reference to Figure 4, at CRF = 22, 90% of undetected persons are encircled by a curve with radius of 112 pixels (after the image has been resized to the input image size of the YOLOv5 network which is 544x544). At CRF = 47, 90% of undetected persons are enclosed by a curve with a radius of 166 pixels, indicating that as compression increases, the minimum object size that we can reliable detect also rises.

### 3.2. Variation in $F_1$ performance in various scenes

The spread of $F_1$ scores for each unique scene (i.e., each of 50 test videos) at every compression level is shown in the boxplot in Figure 5. The $F_1$ scores here refer to the weighted average $F_1$ value, which is calculated based on the $F_1$ scores for each object category weighted by the number of samples in each category, as per Equation 1. $\overline{F}_1(l, v)$ is the weighted average $F_1$ score at a given CRF level, $l$, for a given video, $v$. The $F_1$ score for each individual object category, $c$, at a given CRF level and for a given video, $F_1(l, v, c)$, are weighted by the number of true observations for a given class in a particular video, $N_{v,c}$, divided by the total number of objects in that video $N_v$.

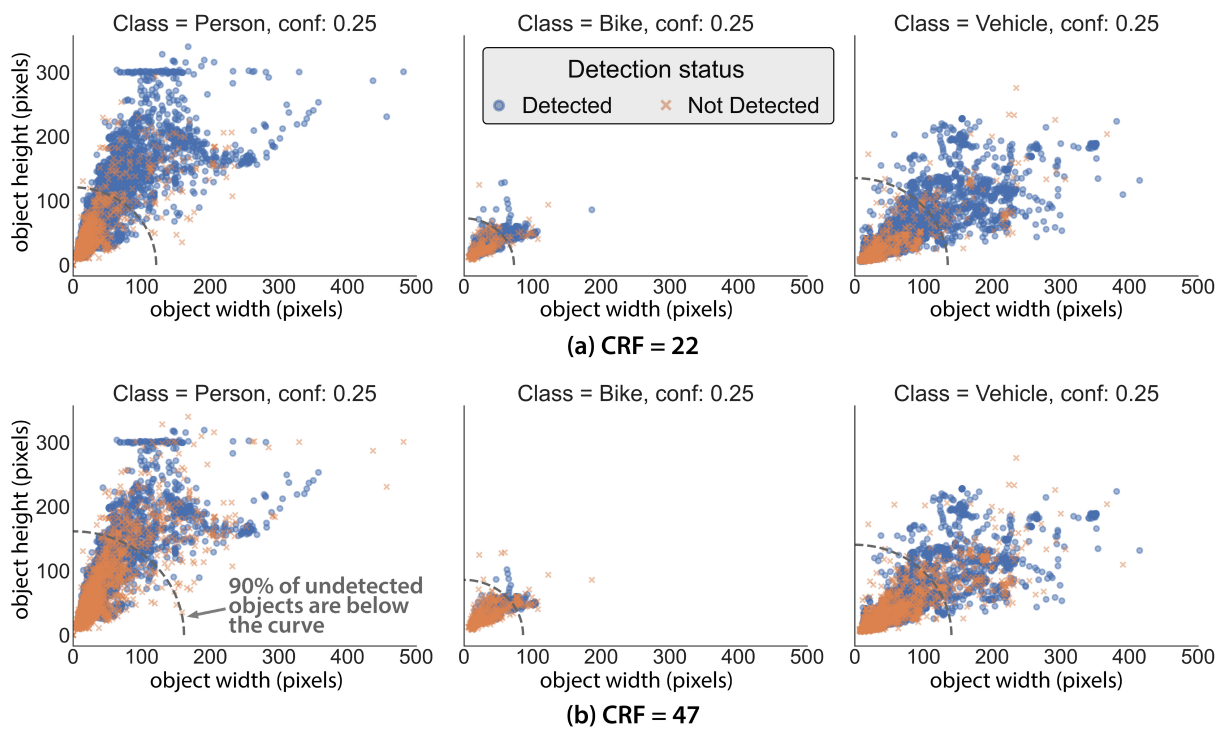

Figure 4: The effect of object size on detection success. Top row (a) shows the variation in detection success with respect to object size for each category at the lowest compression level (CRF = 22). Bottom row (b) shows the same at the highest compression level (CRF = 47). Images are resized (using letterbox method) to match the input size to the YOLOv5 network (544x544 pixels). We can see that YOLOv5 struggles to detect smaller objects, and this is especially evident when the level of compression is high. The curved dashed lines enclose 90% of undetected objects. Larger objects tend to be more resilient to increased levels of compression.

$$\bar{F}_1(l, v) = \sum_c \frac{F_1(l, v, c) \times N_{v,c}}{N_v} \quad (1)$$

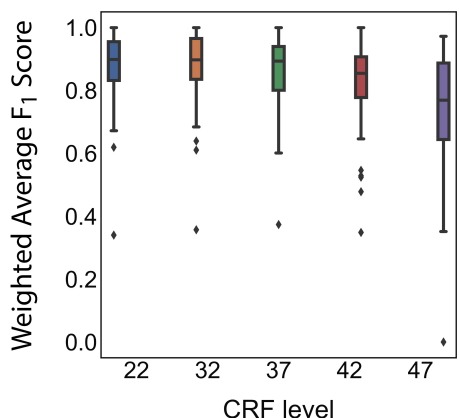

Figure 5: Boxplot showing the variation in $F_1$ scores for test videos at each compression level.

It may be observed that, at lower compression settings (CRF = 22,32,37), the $F_1$ scores are consistent, whereas at the highest compression level (CRF = 47), the $F_1$ scores deviate markedly, with some scenes maintaining decent detection performance while other scenes give rise to poor detection results, such as the scenes shown in Figure 6

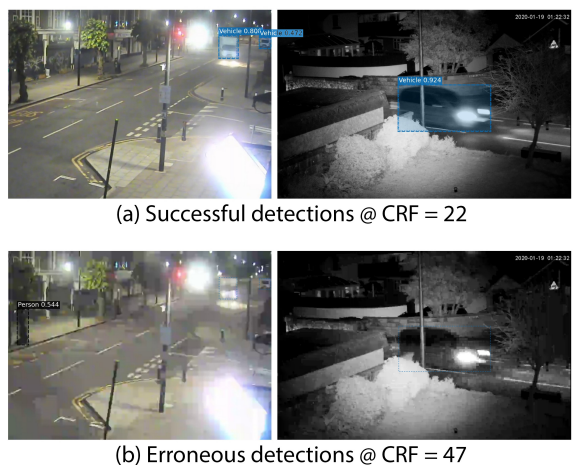

Figure 6: (a) Objects are successfully detected in minimally compressed scenes (CRF = 22), while in (b) there were missed detections in the corresponding compressed scenes (CRF = 47).

Table 1: Average bitrate, video quality scores and object detection performance at each compression level.

| CRF level | Weighted Average Bitrate (Mb/s) | Weighted Average PSNR (dB) | Weighted Average SSIM (dB) | Weighted Average VMAF | Weighted Average $F_1$ |
|---|---|---|---|---|---|
| 22 | 2.32 | 44.86 | 26.78 | 96.58 | 0.89 |
| 32 | 0.63 | 38.16 | 19.24 | 88.19 | 0.90 |
| 37 | 0.34 | 35.06 | 15.96 | 79.38 | 0.89 |
| 42 | 0.19 | 32.06 | 12.92 | 67.49 | 0.84 |
| 47 | 0.12 | 29.05 | 9.95 | 53.16 | 0.76 |

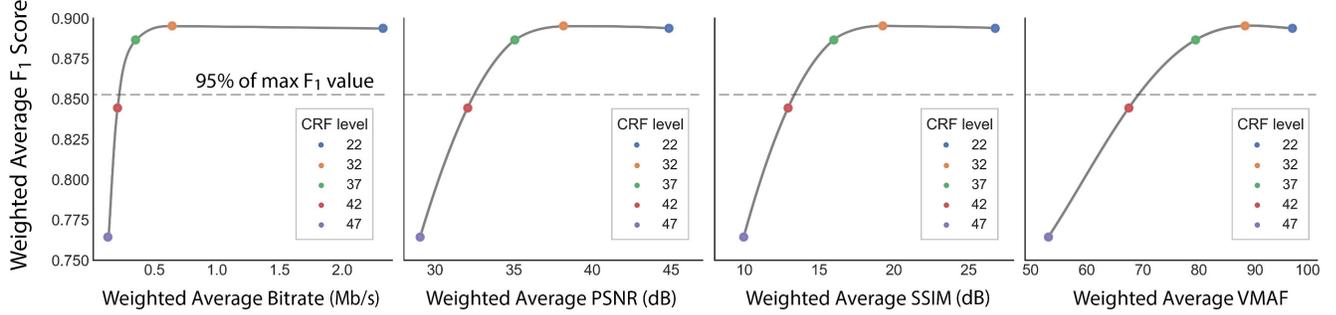

Figure 7: Plot of weighted average $F_1$ scores versus weighted average bitrate, PSNR, SSIM, and VMAF scores.

A closer examination of the scenes that produce bad detection results at high compression settings reveals some interesting insights. Scenes that are characterized by complex lighting conditions, such as dark nighttime scenes and unevenly illuminated scenes, such as the scenes shown in Figure 6, are particularly impacted by the level of compression. The detection of fast-moving objects, such as vehicles, also suffers at high compression levels.

### 3.3. Relationship between detection performance and video quality metrics

The average detection performance at each CRF level versus the corresponding average bitrates and weighted average video quality metrics (PSNR, SSIM, and VMAF scores) are summarized in Table 1 and plotted in Figure 7.

Each video quality metric produces one score per video frame. The weighted average scores at each CRF level are computed by averaging the per-frame scores. Computation of the average VMAF score, $\overline{VMAF}(l)$, at CRF level is presented in Equation 2:

$$\overline{VMAF}(l) = \frac{1}{V} \sum_v \sum_i \frac{VMAF(l,v,i)}{I_v} \quad (2)$$

where $V$ is the total number of test videos ($V = 50$), and $\overline{VMAF}(l,v,i)$ is the VMAF score at CRF level $l$, for the $i^{th}$ extracted frame from the $v^{th}$ video in the test set. $I_v$ is the total frames extracted from the $v^{th}$ video. Weighted average scores for other metrics were calculated in the same fashion.

It may be observed from Figure 7 that there is only a slight drop in the average detection performance when videos are moderately compressed (CRF = 37) compared to when videos undergo the least compression (CRF = 22). When CRF = 37, the average $F_1$ score is 0.887, while at CRF = 22, the average $F_1$ score is 0.894. While there is negligible difference in terms of the detection performance rates at CRF values of 22 and 37, there is, crucially, a significant reduction in the bitrates (2.32 Mb/s vs 0.34 Mb/s). The absolute value of the average bitrates at each CRF level are not overly meaningful since the bitrates are affected by scene activity and resolution, however, the general trend is revealing (i.e., how the bitrates decrease with increasing CRF level). This shows that we can achieve similar rates of detection success using lower bitrate videos, which has important storage and transmission implications.

However, this only holds true up to a point. It may be observed in Figure 7 that the performance of YOLOv5 declines markedly at the higher CRF values. When the CRF = 42, the average $F_1$ score is less than 95% of the peak $F_1$ score, and when the CRF = 47, the detection performance drops to 85% of the peak $F_1$ score. This underscores the value of knowing how detection performance varies with respect to the level of compression so that a suitable cut-off CRF value can be established, beyond which the detection performance becomes unacceptable.

Closer inspection of Table 1 reveals that the best average $F_1$ score is achieved when CRF = 32 (average $F_1$ score = 0.90) rather than at the lowest compression level when CRF = 22 (average $F_1$ score = 0.89). A possible reason for this could be due to the noise-attenuation properties of lossy compression methods such as H.264 which aim to remove unnecessary components of the video, whilst retaining the underlying signal. Removal of such redundant components may have a positive effect on detection performance.

Video quality is appraised at each CRF level using several video quality metrics (PSNR, SSIM, and VMAF). It may be observed in Figure 7 that in the case of the VMAF score, the three CRF levels that all provide similarly good detection performances (CRF = 22,32,37) are clustered quite closely together. This is in contrast with the PSNR and SSIM scores where there is greater spread amongst these CRF levels despite there being minimal difference in detection performance. This suggests that VMAF could be used as a criterion for better encoding decisions and predicting if a video is overly compressed to the point that it will have an appreciably adverse effect on detection performance. For example, based on our experiments, we could stipulate that a compressed video with a VMAF score above 80 would be unlikely to suffer any significant drop in detection performance, although more works needs to be carried out in this direction.

### 3.4. Retraining YOLO with compressed data

In an effort to boost the performance of the object detection system when applied to highly compressed imagery, we retrained the YOLOv5 model with an additional corpus of 22,571 training images that contained realistic video compression artefacts. These images were sourced from a sub-set of MS-COCO [4] images. Video compression artefacts were introduced by: i) converting images into short segment videos, ii) simulating motion by dynamically applying local spatial warping deformations, iii) compressing the video segment using a randomly chosen codec, and iv) extracting the frame from the compressed video segment that will be used as a training image. Examples of some original input images and corresponding corrupted images are shown in Figure 8.

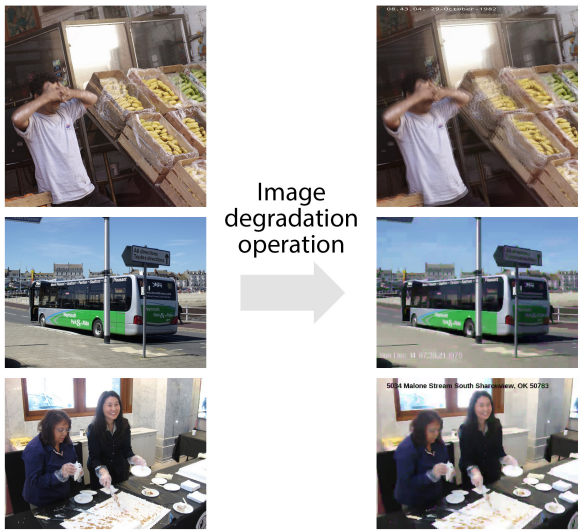

Figure 8: Left column: sample input images, and Right column: degraded images that feature video compression artefacts and other characteristics of surveillance style footage such as timestamps. Inclusion of these degraded images in the training dataset resulted in the YOLOv5 object detector performing slightly better when applied to heavily compressed footage.

Additionally, we observed that the baseline object detection network could not reliably detect objects when a timestamp (or some other text) was overlaid on top of the object of interest. For this reason, we introduced a new custom augmentation that randomly inserted text over images. The text consisted of random dates and addresses. The text was designed to mimic the style of text that is often present in CCTV footage. Various properties of the text are randomized such as the text size, position, font, and color.

The performance of the baseline YOLOv5 model (Model A) and the YOLOv5 model trained on the additional corrupted dataset (Model B) when applied to most heavily compressed test dataset (CRF = 47) is shown in Figure 9.

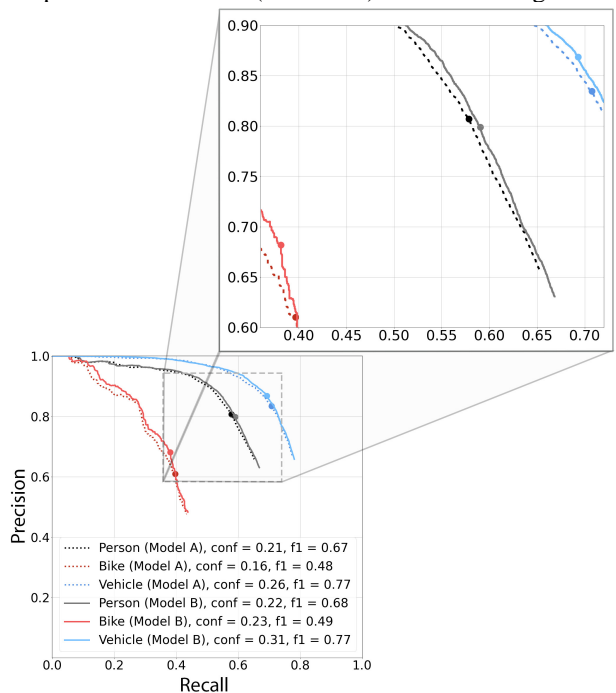

Figure 9: Precision-Recall curves for the baseline YOLOv5 model (Model A) and the YOLOv5 model that was trained with an additional corpus of 22,571 training images that contained realistic video compression artefacts (Model B). The models are applied to the highest compressed test dataset (CRF=47). The inclusion of compressed training images led to a marginal improvement in the $F_1$ score for each class.

The model trained with the additional corrupted training images demonstrated a marginal improvement, (~1% improvement in the $F_1$ scores for each class).

### 4. Discussion and conclusion

It is important for automated surveillance monitoring systems to be able to reliably detect objects and events, even when contending with complex scenes. Designing robust monitoring systems necessitates making informed decisions around the right choice of hardware, system configuration parameters, and algorithms. The video compression rate is a key configuration parameter that is often overlooked in the design of these systems.

In this study, we look at how video compression affects the performance of an object detection network. We train a YOLOv5 model to detect three object categories, namely: persons, bikes, and vehicles. The detection performance is generally robust to moderate levels of compression; we found that there is a negligible difference in the detection performance when videos are minimally compressed (CRF = 22) versus when they are moderately compressed (CRF = 37). This is noteworthy as we can realize significant bitrate savings without incurring any penalty in terms of detection performance – the average bitrate for videos encoded at CRF = 22 was 2.32 Mb/s, while it was almost 7 times lower for videos encoded at CRF = 37, at 0.34 Mb/s.

We point out situations where parsimonious bitrates are not recommended. This includes cases where fast moving objects travel through dark scenes or scenes with uneven illumination (i.e., scenes with both over-exposed and poorly lit regions). In these complex scenes, videos that are heavily compressed start to breakdown and have low visual quality.

In practical terms, the best measure of video quality is the success rate of computer vision algorithms such as object detection models. As part of this study, we examined how the detection performances correlate with established video quality metrics including PSNR and SSIM at a given CRF rate. In recent years, emphasis has been put on developing various methods and techniques for evaluating the perceived quality of video content by human observers. A popular perceptual metric that has emerged is VMAF. While VMAF has principally been geared towards evaluating media for the entertainment sector as opposed to for CCTV tasks, we found that it may be a good proxy for predicting if a video is overly compressed to the point that it appreciably harms detection performance. Human observers typically interpret videos with a VMAF score of 70 as a vote between "good" and "fair". We found that, on average, videos with a VMAF score above 70 achieved good detection performances (as measured by $F_1$ score). More work needs to be carried to this end to establish if there is a solid basis for linking VMAF score and detection success. A potentially promising future research direction involves quantifying the video quality degradations using a video quality metric such as VMAF so that we can control the quality of the video data, and consequently, ensure detection success is not hampered by low-quality, excessively compressed input video data.

Most available open-source datasets contain images that are not representative of surveillance style footage, especially heavily compressed surveillance style footage. Training deep neural networks using these images alone will likely produce a model that is not well-suited for processing real-world surveillance footage. In order to address this, we trained a new YOLOv5 model using an additional 22,571 images containing realistic video compression artefacts and characteristics of CCTV footage. This led to a 1% improvement in $F_1$ score for most classes when applied to the highest compressed imagery (CRF = 47). Although this is only a slight improvement, there is scope to train with more corrupted images and adopt more sophisticated augmentation operations to improve detection success. This is left as another possible research direction.

## Acknowledgements

The authors acknowledge funding from the Disruptive Technologies Innovation Fund, Project: 166687/RR.